\documentclass[runningheads]{llncs}
\usepackage[T1]{fontenc}
\usepackage{amsfonts}
\usepackage{amsmath}
\usepackage{subcaption}
\usepackage{hyperref}
\usepackage{graphicx,verbatim}
\begin{document}
\title{BiasICL: In-Context Learning and Demographic Biases of Vision Language Models}
\titlerunning{BiasICL: In-Context Learning and Demographic Biases}

\author{Sonnet Xu\textsuperscript{*}\inst{1}\thanks{Corresponding author: sonnet@stanford.edu} \and
Joseph D. Janizek\textsuperscript{*}\inst{1,2} \and
Yixing Jiang\inst{1} \and
Roxana Daneshjou\inst{1}}
\authorrunning{S. Xu et al.}
\institute{
    Stanford University, Stanford, CA 
    % \and Springer Heidelberg, Tiergartenstr. 17, 69121 Heidelberg, Germany
    % \email{lncs@springer.com}\\
    % \url{http://www.springer.com/gp/computer-science/lncs} 
    \and
    Virginia Mason Internal Medicine Residency, Seattle, WA\\
    % \email{\{abc,lncs\}@uni-heidelberg.de}
    \textsuperscript{*}Equal Contribution
}

% \author{Anonymized Authors}  %% Added for anonymized MICCAI 2025 submission
% \authorrunning{Anonymized Author et al.}
% \institute{Anonymized Affiliations \\
%     \email{email@anonymized.com}}

\maketitle              % typeset the header of the contribution
\begin{abstract}
Vision language models (VLMs) show promise in medical diagnosis, but their performance across demographic subgroups when using in-context learning (ICL) remains poorly understood. We examine how the demographic composition of demonstration examples affects VLM performance in two medical imaging tasks: skin lesion malignancy prediction and pneumothorax detection from chest radiographs. Our analysis reveals that ICL influences model predictions through multiple mechanisms: (1) ICL allows VLMs to learn subgroup-specific disease base rates from prompts and (2) ICL leads VLMs to make predictions that perform differently across demographic groups, even after controlling for subgroup-specific disease base rates. Our empirical results inform best-practices for prompting current VLMs (specifically examining demographic subgroup performance, and matching base rates of labels to target distribution at a bulk level and within subgroups), while also suggesting next steps for improving our theoretical understanding of these models.
\url{https://github.com/DaneshjouLab/BiasICL}
% \url{https://anonymous.4open.science/r/MICCAI2025-7B4D/README.md}
% We introduce a subgroup-aware classification method that improves VLM diagnostic accuracy across demographic subgroups. https://anonymous.4open.science/r/MICCAI2025-7B4D/

% Authors must provide keywords and are not allowed to remove this Keyword section.

\end{abstract}
\section{Introduction and background}

% In-context learning (ICL), or the capability of models to ``learn'' from demonstrations within prompts, is an exciting new phenomenon in the era of large language models \cite{DBLP:journals/corr/abs-2005-14165}. In the area of medical imaging in particular, ICL is particularly exciting. Medical datasets are costly to produce due to the extent of privacy and regulations protecting the data, and the high-degree of professional human expertise frequently required to create ground truth labels. In contrast to the tens or hundreds of thousands of examples required to create accurate traditional supervised deep learning models, ICL opens the possibility of adapting a model to a particular task with a significantly smaller handful of demonstrating examples. ICL has been shown to be successful for text-based medical tasks \cite{nori2023can}, and more recently has shown promising results for multimodal vision language models (VLMs) prompted to complete computer vision tasks like cancer pathology image classification \cite{ferber2024context}, chest radiograph classification, and dermatology image classification \cite{jiang2024many}.

In-context learning (ICL), or the capacity of large language models (LLMs) to adapt to new tasks from a handful of demonstrations in the prompt, has emerged as an exciting development in AI \cite{DBLP:journals/corr/abs-2005-14165}. For medical AI tasks, this technique circumvents the large datasets required for supervised learning, which are costly due to privacy regulations and the clinical expertise required for data annotation. While traditional deep learning often demands tens or hundreds of thousands of labeled samples, ICL enables models to be rapidly customized to a new task using only a few examples. Although first demonstrated predominantly in text-based medical tasks \cite{nori2023can}, recent work has shown that ICL also improves the performance of vision-language models (VLMs) for tasks such as cancer pathology image classification \cite{ferber2024context}, chest radiograph classification, and dermatology image classification \cite{jiang2024many}. 

Despite its promise, ICL remains imperfectly understood and can lead to high variance in predictive accuracy based on seemingly minor modifications in prompts. In text-only settings, prior research demonstrates that LLMs may over-predict labels that appear more frequently in the prompt, appear last in the prompt, or are simply more common in their pre-training data \cite{zhao2021calibrate}. 

Prior to deployment in medical settings, VLM safety and bias need to be better understood. 
Substantial work has been devoted to understanding whether machine learning models have problems with demographic biases or fairness issues. This includes work on previous generations of supervised deep learning vision models \cite{seyyed2020chexclusion}, text-based large language models \cite{omiye2023large}, and now even vision-based large multi-modal VLMs \cite{yang2024demographic,wu2024fmbench}. 

Our work differs from prior bias and fairness work in the following way -- we specifically aim to understand how prompting VLMs using ICL impacts demographic fairness, as opposed to prior work that investigates text models or investigates multi-modal models in the 0-shot setting. Consequently, we focus on large, commercial API-based VLMs, with the capacity to handle interleaved images and text with sufficiently large contexts to handle many demonstrations, in line with prior work from Jiang et al. \cite{jiang2024many}. 

% Concurrently, there has been an extensive study of demographic biases in \emph{supervised} deep learning models for medical imaging, showing that these models can learn patient attributes such as race, sex, and age \cite{yi2021radiology,gichoya2022ai,munk2021assessment}. 
% % sometimes using easy-to-learn ``shortcuts'' like laterality markers or imaging projections \cite{zech2018variable,degrave2021ai,janizek2020adversarial}. 
% These two lines of work (bias in ICL for text-based tasks and demographic bias in supervised medical image models) raise a natural question: how do prompt-based VLMs handle demographic subgroups, and can ICL inadvertently propagate or exacerbate biases?

This paper investigates how ICL with VLMs may inadvertently shift predictive distributions in ways that influence demographic fairness. First, we demonstrate that VLMs show a ``majority label bias,'' similar to what has been observed in LLMs. Second, we find that VLMs exhibit a \emph{demographic group} majority label bias, indicating sensitivity to base rates not just overall but also within specific demographic subgroups. This sensitivity seems to depend on how accurately the models can identify the demographic subgroups. Finally, we show that even after ensuring subgroup balance in prompts, ICL can increase the level of bias in predictions. Surprisingly, in this setting, ICL can lead to improvement in accuracy of prediction of one subgroup directly to the detriment of accuracy in another.

\section{Methods}
\subsection{Datasets}
% \subsubsection{Diverse Dermatology Images (DDI)}

\begin{figure}
    \centering    
        
        \includegraphics[width=\textwidth]{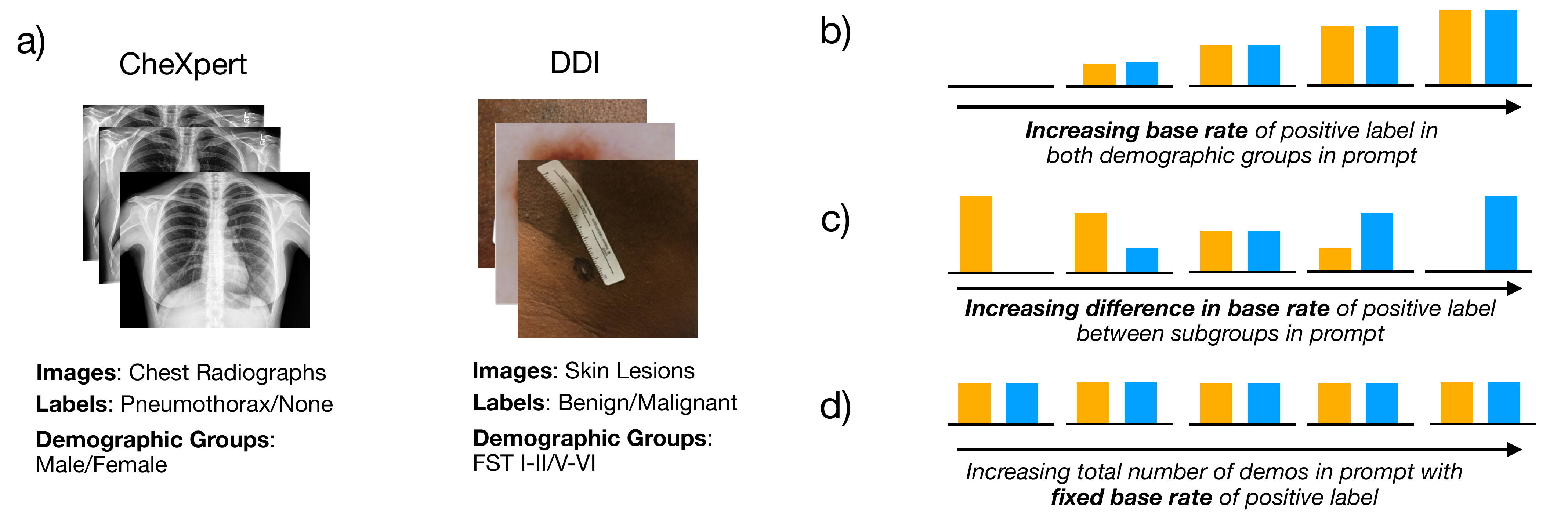}
   
    \caption{\textbf{Overview}. CheXpert and DDI \textbf{(a)} were used to investigate a variety of different biases, including: \textbf{(b)} Majority label bias, or the tendency of models to predict more prevalent labels in the prompt more frequently; \textbf{(c)} a new bias introduced in our paper called group majority label bias, or the tendency of models to be swayed by the majority label seen using ICL \textit{within a particular demographic subgroup} when encountering test examples from that same subgroup; and \textbf{(d)} ICL bias, or the extent to which models learn disparities between groups as the number of demos in a prompt increases. In (b-d), orange and blue bars represent different demographic groups, and the height of each bar represents the fraction of positive labels in the prompt within that subgroup.}
    \label{fig:concept}
\end{figure}
% The Diverse Dermatology Images (DDI) dataset consists of clinical images of skin lesions paired with labels of biopsy-proven diagnoses (benign vs malignant) for these lesions, as well as labels of the Fitzpatrick skin type (FST) of the patients in the images \cite{daneshjou2022disparities}. Fitzpatrick skin type is a classification scheme for skin tone, and the DDI dataset contains 208 images (159 benign and 49 malignant) of FST I–II patients (light skin tones), and 207 images (159 benign and 48 malignant) of FST V–VI patients (dark skin tones). DDI also includes images of patients with intermediate skin tones (FST III–IV), but in line with Daneshjou et al. \cite{daneshjou2022disparities}, we focus our analysis on FST I–II and FST V–VI to better assess model performance across more distinct skin types (see Fig. \ref{fig:concept}a). We split this dataset into two subsets, a demo set (311 patients) for use in prompting and a test set (104 patients) for use in evaluation. We maintain an evenP demographic and diagnosis balance across the demo and test sets, such that each set is comprised of half FST I–II patients and half FST V-VI patients and such that the base rate of maligancy in both FST I–II and FST V-VI patients in both the demonstration and test sets is approximately 25 percent.
The Diverse Dermatology Images (DDI) dataset contains clinical images of skin lesions with biopsy-proven benign or malignant labels and Fitzpatrick skin type (FST) labels \cite{daneshjou2022disparities}. Following Daneshjou et al. \cite{daneshjou2022disparities}, we focus on FST I–II (light skin tones) and FST V–VI (dark skin tones), comprising 208 (159 benign, 49 malignant) and 207 (159 benign, 48 malignant) images, respectively (see Fig. \ref{fig:concept}a). We exclude FST III–IV images to highlight performance across more distinct skin tones. We split these into a 311-image demo set and a 104-image test set, ensuring equal representation of FST I–II and V–VI and maintaining a balanced \~25\% malignancy rate in each group.

CheXpert is a large dataset of 224,316 chest radiographs from 65,240 patients, with 14 labels automatically extracted from radiology reports \cite{irvin2019chexpert}. Because we focus on demographic differences in prompt demonstrations, we limit our analysis to a small subset: 400 patients in a demo set and 100 in a test set, each split evenly by sex. In each demographic subgroup (in both demo and test sets), half the radiographs are labeled “pneumothorax,” and half are not (see Fig. \ref{fig:concept}a).

% The CheXpert dataset is a large collection of 224,316 chest radiographs of 65,240 patients with 14 labels automatically extracted from associated radiology reports using an automated labeler \cite{irvin2019chexpert}. Because our study tries to focus on how demographic differences in prompt demonstrations impact model performance, we restrict our focus to a small subset of this dataset. We create a demo set of 400 patients and a test set of 100 patients where half of the patients are male and half of the patients are female. In each demographic subgroup in both the demo and test sets, half of the radiographs are positive for the ``Pneumothorax'' label and half of the radiographs are negative (see Fig. \ref{fig:concept}a).

\subsection{Models}
Our study focuses on API-based, commercial VLMs. We investigate three models from three different providers: GPT-4o (with the specific endpoint ``gpt-4o-2024-05-13''), Gemini 1.5 Pro (with the specific endpoint ``gemini-1.5-pro-preview-0409''), and Claude3.5-Sonnet (with the specific endpoint ``claude-3-5-sonnet-20241022''). We use the API service provided by OpenAI for GPT-4o, the API service provided by Google Cloud on Vertex AI for Gemini 1.5 Pro, and the API service provided by Anthropic for Claude3.5-Sonnet. We set the temperature to zero for all models and a random seed for GPT-4o to obtain more deterministic responses. These models were selected because they all have large contexts, high accuracy across many multi-modal benchmark tasks, and most importantly, have the capacity in their context to handle many interleaved images and texts.

\subsection{Prompting and evaluation}

LLMs and VLMs can be sensitive to prompting and evaluation strategies \cite{biderman2024lessons}, so we provide the exact prompts in our 
\href{https://github.com/DaneshjouLab/BiasICL}{GitHub}.
% \href{https://anonymous.4open.science/r/MICCAI2025-7B4D/README.md}{Anonymized Repository}
Following Jiang et al. \cite{jiang2024many}, our prompts include: (1) a preamble specifying response format; (2) demonstration examples (image, question, possible answers, correct answer); and (3) test images/questions. We also use Batch Querying \cite{jiang2024many}, a method explored in a variety of prior works that leads to more efficient and cheaper inference by batching multiple test questions together in a single prompt.

To evaluate models, we parse answers from text completions because many API-based models do not provide logprob access \cite{mirza2024large,laurent2024lab}. When formatting issues arose, we simply re-sent queries, as in Jiang et al. \cite{jiang2024many}, rather than using an LLM fallback parser \cite{mirza2024large}.

% A well-known problem in evaluating LLMs and VLMs is sensitivity to prompting and parsing/evaluation methods \cite{biderman2024lessons}. We therefore include the exact prompts used in each experiment in our \href{https://anonymous.4open.science/r/MICCAI2025-7B4D/README.md}{Anonymized Repository}. To briefly summarize the prompts used, we closely followed the approach used in recent, state-of-the-art work on in-context learning for VLMs \cite{jiang2024many}. Our prompts included the following components: (1) a preamble, instructing the model of the generally expected format of response; (2) a set of demonstrations sampled from our demo sets, where each demonstration includes an image, a question, the possible answer choices, and the correct answer for each image; and (3) the test images and questions for which we want the model to make predictions. Following Jiang et al. \cite{jiang2024many}, we use Batch Querying, a method explored in a variety of prior works that leads to more efficient and cheaper inference by batching multiple test questions together in a single prompt.

% To evaluate models, we directly parse answers from the text completions. This approach is necessary because not all of the large, API-based models provide direct access to the output token logprobs \cite{mirza2024large,laurent2024lab}. When we encountered parsing errors due to models not following formatting instructions, rather than using an LLM as a fall-back parser as in Mirza et al. \cite{mirza2024large}, we followed the approach in Jiang et al. \cite{jiang2024many} and simply re-sent the questions to the API.

\subsection{Bias measures}

We considered three different types of bias when assessing in-context learning (ICL). The first, \textbf{majority label bias}, first observed by Zhao et al. \cite{zhao2021calibrate}, measures how sensitive models are to the frequency of positive labels in the demonstration set (see Fig. \ref{fig:concept}b). Specifically, we measure how the average prediction made by the model over all images in the test set, $\mathbb{E}_{(x,y) \sim \mathcal{D}_{\text{test}}}[f(x)]$, is impacted by varying the proportion of positive labels in the prompt, $\mathbb{E}_{(x,y) \sim \mathcal{D}_{\text{demo}}}[y]$.We assess this bias since recent work has suggested that newer models may be more robust to this bias \cite{gupta2023robust}.

The second, which we termed \textbf{group majority label bias}, measures how sensitive models are to the frequency of positive labels within \textit{each demographic group} in the demonstration set (see Fig. \ref{fig:concept}c). Specifically, we measure how the difference in the average prediction made by the model over all images in the test set conditional on subgroup membership ($\mathbb{E}_{(x,y) \sim \mathcal{D}_{\text{test}}}[f(x) \mid g(x) = 1] - \mathbb{E}_{(x,y) \sim \mathcal{D}_{\text{test}}}[f(x) \mid g(x) = 0]$) is impacted by varying the difference in the proportion of positive labels between demographic groups in the prompt ($\mathbb{E}_{(x,y) \sim \mathcal{D}_{\text{demo}}}[y \mid g(x) = 1] - \mathbb{E}_{(x,y) \sim \mathcal{D}_{\text{demo}}}[y \mid g(x) = 0]$).

Finally, to isolate the effects of number of demonstrations in the prompt, we fix the base rate of positive labels in the prompt equal to the base rate in the test set, and fix the base rates equal across both demographic subgroups. Then we measure how the difference in the average prediction made by the model over all images in the test set conditional on subgroup membership ($\mathbb{E}_{(x,y) \sim \mathcal{D}_{\text{test}}}[f(x) \mid g(x) = 1] - \mathbb{E}_{(x,y) \sim \mathcal{D}_{\text{test}}}[f(x) \mid g(x) = 0]$) is impacted by increasing the number of demonstrations in the prompt. We refer to this as \textbf{ICL bias} (see Fig. \ref{fig:concept}d). We also examined the difference in the predictive performance (measured by F1 score) between demographic groups as the number of prompt demonstrations are increased.

\section{Results}

\subsection{VLMs learn a majority label bias}

\begin{figure}
    \centering
    \includegraphics[width=0.80\textwidth]{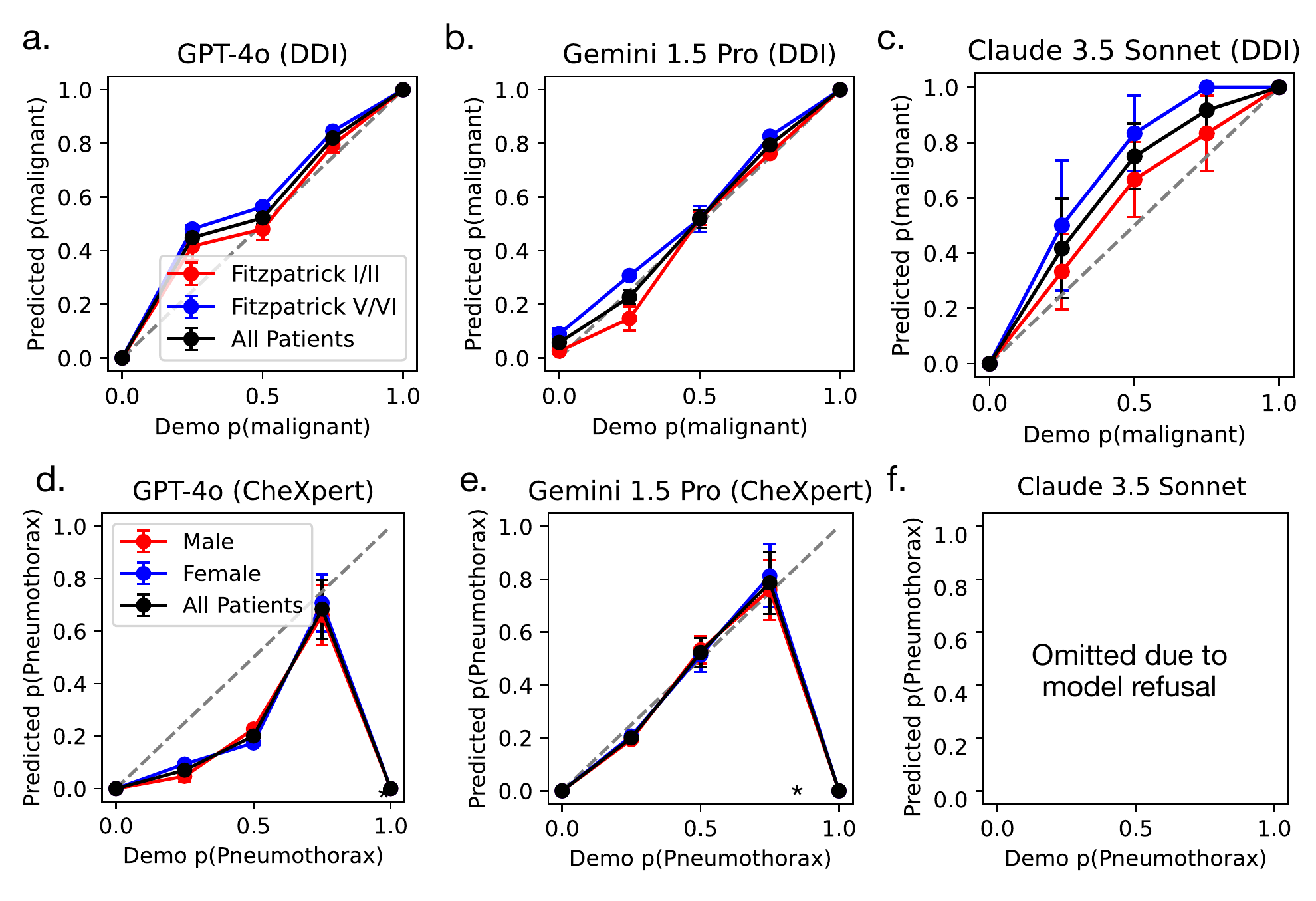}
        
    \caption{Majority label bias -- models more frequently predict labels that are more frequent in the prompt. (a-c) Prediction of malignancy on the DDI dataset. (d-f) Prediction of pneumothorax on the CheXpert dataset. Error bars = standard error over three independent runs with different random seeds for selection of demonstration examples from the dataset and ordering of demonstrations in the prompts.}
    % Asterisks in d and e mark two outlier points. f depicts that results for Claude 3.5 Sonnet could not be obtained in this setting due to numerous model refusals}
    \label{fig:majority_bias}
\end{figure}

When we prompted models with a constant number of demonstrations, but increased the frequency of demonstrating examples with positive labels, we see that the models' outputs become biased towards that prediction (see Fig. \ref{fig:majority_bias}). While this relationship appears to mostly be nearly linear, we observe two outlier points in the CheXpert dataset in the two cases when all demonstrating examples are positive. We also omitted Claude 3.5 Sonnet results on the CheXpert dataset in this experiment, as the model tended to abstain too frequently to obtain reliable results.

\subsection{VLMs learn a demographic group majority label bias}

\begin{figure}
    \centering
    \includegraphics[width=0.95\textwidth]{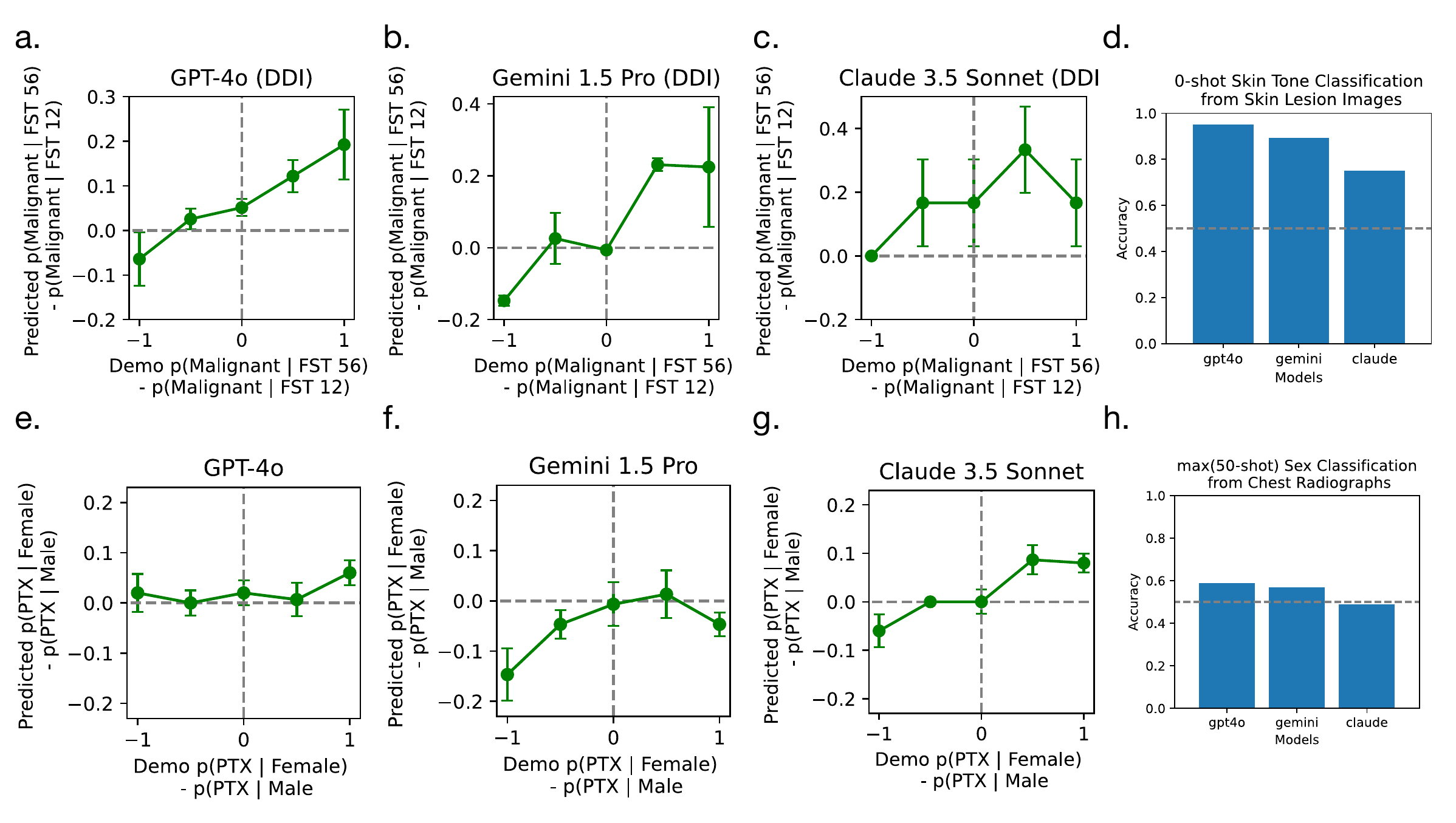}
        
    \caption{Demographic group majority label bias. (a-c) Malignancy prediction on DDI dataset; (e-g) Pneumothorax prediction on CheXpert. Error bars = standard error over three independent runs with different random seeds for demonstration selection and prompt ordering. (d) 0-shot accuracy for patient Fitzpatrick skin type prediction from dermatology images; (h) maximum 0-to-50-shot accuracy for patient sex prediction from chest radiographs.}
    % Demographic group majority label bias across 3 models. (a-c) show results for GPT-4o, Gemini 1.5 Pro, and Claude 3.5 Sonnet for the prediction of malignancy on the DDI dataset, while (e-g) show results for the same models for the prediction of pneumothorax on the CheXpert dataset. Error bars represent the standard error over three independently repeated runs with different random seeds for selection of demonstration examples from the dataset and for ordering of demonstrations in the prompts. (d) Shows the 0-shot accuracy of the three models at predicting patient Fitzpatrick skin type from dermatology images, while (h) shows the maximum of 0- to 50-shot accuracy of the three models at predicting patient sex from chest radiographs.
    
    \label{fig:group_bias}
\end{figure}

After demonstrating that VLMs are sensitive to the overall base rate of label frequency in their prompts, we wanted to investigate whether it is important to pay attention to the base rate of different labels \textit{within} different subgroups. We refer to this property as a demographic group majority label bias. In Fig. \ref{fig:group_bias}, we hold the total number of samples in the prompt constant and increase the \textit{difference} between the base rate of positive labels in the two demographic subgroups. Across most model-dataset pairs, we see that models do learn this bias from the prompt. We notice that this effect is more pronounced in the DDI dataset, where the maximum difference in subgroup mean prediction is 30\% per model tested (Fig. \ref{fig:group_bias}a-c), compared to the CheXpert dataset, where the maximum difference is closer to 10\% (Fig. \ref{fig:group_bias}e-g). 

We hypothesized that this might be due to differences in the ability of models to detect and predict those demographic subgroups in the first place, particularly as there has been extensive study of demographic biases in \emph{supervised} deep learning models for medical imaging, showing that these models can learn patient attributes such as race, sex, and age \cite{yi2021radiology,gichoya2022ai,munk2021assessment}. Across models tested, in the 0-shot setting, models can highly accurately classify patients' Fitzpatrick skin type (see Fig \ref{fig:group_bias}d). We also investigate the ability of models to identify demographic subgroups from chest radiographs, namely, patients' sex. Because predictive performance was low in the 0-shot setting, we increased the number of demonstrating examples in this experiment up to 50, and plotted the max accuracy per model over that range. We found that while models could predict sex with greater-than-random accuracy (see Fig \ref{fig:group_bias}h), the accuracy was generally much lower than for skin tone prediction. 

Finally, given that previous work had demonstrated that supervised, deep convolutional neural network models could predict patients' self-reported race from chest radiographs with very high accuracy \cite{gichoya2022ai}, we also probed the ability of VLMs to make this prediction. Despite substantial efforts to re-engineer prompts, all three VLMs invariably refused to attempt to make this classification.

These findings suggest that VLMs are less capable of identifying demographic subgroup identities than previous generations of supervised deep learning models, and also that as their capability in this area grows, their susceptibility to this particular group majority label bias will increase as well. This is in keeping with prior work suggesting that supervised learning models may have a tendency to use demographic groups as ``shortcuts'' or proxies in their predictions \cite{banerjee2023shortcuts,zech2018variable,degrave2021ai,janizek2020adversarial}.

% \subsection{Independently of (S)MLB, ICL can worsen subgroup accuracy}

% After demonstrating how the base rate of labels included in the prompt could bias the model towards different predictions for each demographic subgroup, we wanted to investigate how 

\subsection{ICL alone can increase demographic subgroup bias in VLMs}

\begin{figure}
    \centering
    % First row: GPT-4 and Gemini DDI
    \begin{subfigure}[b]{0.34\textwidth}
        \centering
        \includegraphics[width=\textwidth]{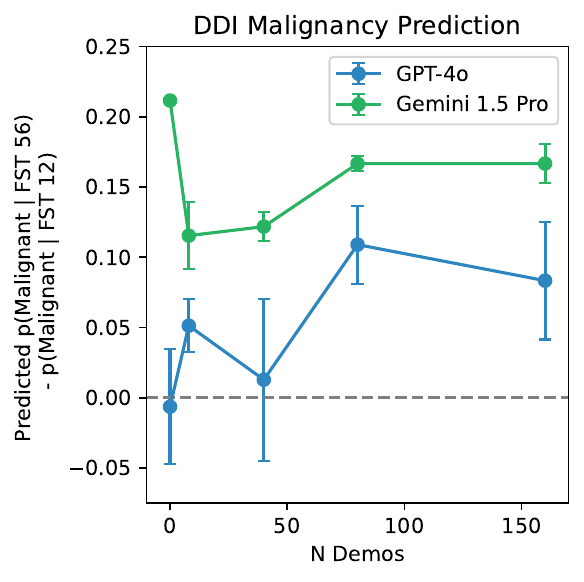}
        % \caption{Bias in Malignancy prediction in DDI dataset}
        % \label{fig:gpt4-ddi}
    \end{subfigure}
    % \hfill
    \begin{subfigure}[b]{0.34\textwidth}
        \centering
        \includegraphics[width=\textwidth]{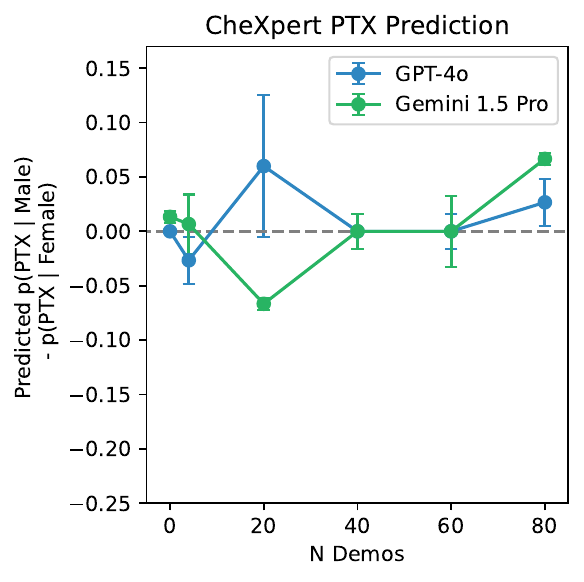}
        % \caption{Bias in PTX prediction in CheXpert dataset}
        % \label{fig:gemini-ddi}
    \end{subfigure}
    \caption{The impact of ICL on the difference between models' average predictions across subgroups when the base rate of positive labels is set equal between subgroups in the prompt.}
    \label{fig:model-comparison}
\end{figure}

We initially hypothesized that adding demos from patients across different demographic subgroups to the prompt with ICL might be able to decrease models' inherent bias. This was because prior work on supervised learning models had shown that fine-tuning using more diverse data could improve subgroup performance \cite{daneshjou2022disparities}. However, our analysis showed that this was not necessarily the case.

For the task of malignant skin lesion prediction on the DDI dataset, GPT-4o demonstrated minimal bias in its predictions across demographic subgroups in the 0-shot setting. However, when provided with additional in-context examples, even those with balanced malignancy rates, the model developed a systematic bias toward predicting higher malignancy rates in patients with Fitzpatrick Skin Types V/VI (roughly 10\%, see Fig. \ref{fig:model-comparison}, left). In contrast, Gemini 1.5 Pro exhibited strong 0-shot bias toward predicting malignancy in FST V/VI patients (around 20\%), and while this bias persisted with the addition of balanced in-context examples, it showed modest attenuation. The models' behavior differed substantially in the CheXpert pneumothorax prediction task (see Fig. \ref{fig:model-comparison}, right). Both GPT-4o and Gemini 1.5 Pro maintained relatively unbiased predictions across gender, independent of number of demonstrating examples. Claude 3.5 Sonnet results were again excluded from this experiment as refusals were too numerous to have a reliable sample size.

\begin{figure}
    \centering
    \includegraphics[width=\textwidth]{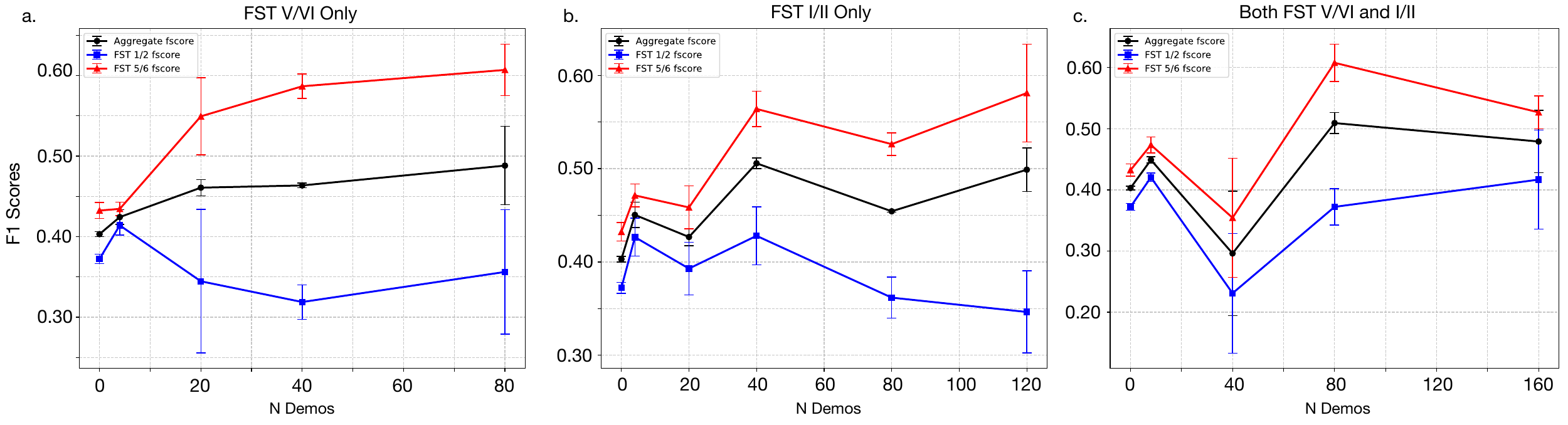}
        
    \caption{The impact of ICL on GPT-4o's predictive performance on the DDI dataset when (a) adding only FST V/VI demos, (b) adding only FST I/II demos, and (c) adding equal numbers of both.}
    \label{fig:ddi_acc}
\end{figure}

In addition to considering bias in terms of the difference in average predictions between groups, we also looked at the \textit{predictive performance} of models across subgroups. Our most interesting results were on the DDI dataset with the GPT-4o model (see Fig. \ref{fig:ddi_acc}). In this experiment, the base rate of malignancy in the prompt was fixed equal to the base rate of malignancy in the test set, and the number of demonstrating examples were increased. Independently of whether the demonstrating examples in the prompt were (a) all of skin type FST V/VI, (b) all of skin type FST I/II, or (c) even numbers of both both, ICL significantly increased the predictive performance (as measured by F1 score) for FSV V/VI patients at the expense of predictive performance for FST I/II patients.

\section{Discussion and limitations}

This empirical work has several immediately-relevant implications for prompting. For medical vision systems that are task-adapted using ICL, developers must consider not only the overall base rate of the labels in the prompt, but also the demographic subgroup-specific base rate of the labels in the prompt. Additionally, even after carefully controlling the base rates of labels per demographic subgroups in prompts, ICL can still lead to exacerbation of differences in predictions across demographic subgroups. This highlights the ongoing importance of lessons learned from the supervised learning era -- hidden stratification, or the tendency of models to have poor performance on important subsets of a population \cite{oakden2020hidden}, is still a relevant property of VLMs ``trained'' with ICL. Consequently, developers should evaluate their models' performance stratified on different relevant subgroups.

Limitations of our current work also suggest future directions for research. While our work shows that ICL with large state-of-the-art API-based models can lead to exacerbated bias between subgroups, it does not uncover the mechanism of \textit{why} this occurs. Future work with open source models, for which the weights and pre-training data can be directly accessed, will allow us to run ablation experiments to determine what aspects of a models training corpus, training process, and architecture primarily contribute to the observed phenomena. Finally, we acknowledge the limitations of our work in evaluating the full social/societal impacts of the \textit{fairness} of these models. We measure bias between subgroups as a quantifiable and mathematical property of these models and datasets. However, a full analysis of the impact of these models would require further investigation of the real sociotechnical contexts in which they might be used \cite{diao2024implications}.

\bibliographystyle{splncs04}
\bibliography{mybib}

\end{document}